\documentclass[journal]{IEEEtran}
\ifCLASSINFOpdf
\else
\fi
\usepackage{amssymb}
\usepackage{amsmath}
\usepackage[ruled,vlined]{algorithm2e}
\usepackage{amsmath,epsfig, booktabs,cite}
\usepackage{subfigure}
\usepackage{xcolor}

\begin{document}
%
\title{
Semi-supervised Hyperspectral Image Classification with Graph Clustering Convolutional Networks
}
%
%
%

\author{
Hao Zeng,
Qingjie Liu,~\IEEEmembership{Member,~IEEE,} 
Mingming Zhang,
Xiaoqing Han,
Yunhong Wang,~\IEEEmembership{Fellow,~IEEE}
\thanks{Hao Zeng, Qingjie Liu (Corresponding author), Minging Zhang and Yunhong Wang are with the State Key Laboratory of Virtual Reality Technology and Systems, Beihang
University, Xueyuan Road, Haidian District, Beijing, 100191, China, and
Hangzhou Innovation Institute, Beihang University, Hangzhou, 310051,China
(email: zy1906915@buaa.edu.cn; qingjie.liu@buaa.edu.cn; sara\_@buaa.edu.cn; 
yhwang@buaa.edu.cn).}
\thanks{Xiaoqing Han is with the Beijing Research Institute of Uranium Geology, Beijing 100029, China (email: 1613739468@qq.com).}}

%
%

\markboth{IEEE Transactions on Geoscience and Remote Sensing}%
{Shell \MakeLowercase{\textit{et al.}}: Bare Demo of IEEEtran.cls for IEEE Journals}
%



\maketitle

\begin{abstract}
Hyperspectral image classification (HIC) is an important but challenging task, and a problem that limits the algorithmic development in this field is that the ground truths of hyperspectral image (HSIs) are extremely hard to obtain.
Recently a handful of HIC methods are developed based on the graph convolution networks (GCNs), which effectively relieves the scarcity of labeled data for deep learning based HIC methods. 
To further lift the classification performance, in this work we propose a graph convolution network (GCN) based framework for HSI classification that uses two clustering operations to better exploit multi-hop node correlations and also effectively reduce graph size.
In particular, we first cluster the pixels with similar spectral features into a superpixel, and build the graph based on the superpixels of the input HSI.
Then instead of performing convolution over this suerperpixel graph, we further partition it into several sub-graphs by pruning the edges with weak weights, so as to strengthen the correlations of nodes with high similarity.
This second round of clustering also further reduces the graph size, thus reducing computation burden of graph convolution.
Experimental results on three widely used benchmark datasets well prove the effectiveness of our proposed framework.
\end{abstract}


%
\IEEEpeerreviewmaketitle

\section{Introduction}
%
%
%
%
\IEEEPARstart{H}{yperspectral} images (HSIs) usually consist of hundreds of spectral bands. 
Such redundant information yields more discriminating feature representations, and on the other hand poses a great challenge to classification. 
Hyperspectral image classification (HIC) has been a long researched task with wide applications such as weather forecast \cite{bloch2019near}, disaster prevention~\cite{zhang2018assessment} and mineral exploration \cite{schmidt2014minerals}.

Early works exploit the spectral curves of different ground objects, mainly including kernel classifier based methods \cite{mercier2003support}, \cite{melgani2004classification} and feature representations based ones \cite{chen2010classification}, and pay little attention to the spatial information in HSIs.
Their performance is often limited due to the Hughes phenomenon (\emph{i.e.} classification accuracy increases gradually with more spectral bands or dimensions, but decreases sharply when the band number reaches some value) as well as spectral noise like light intensity and shadow.
Later, several spatial-spectral joint classification methods have been proposed \cite{li2012spectral, chen2014spectral,  ghamisi2013spectral}.
Among them, Markov random field (MRF) is a popular one, which exploits the strong dependencies between neighboring pixels for classification. 
However, mining such dependencies often demands setting parameters by prior knowledge, and may harm performance when the spectral information of adjacent pixels is diverse and complex.
Also, the handcrafted spectral-spatial features are often not sufficiently representative.

In recent years, the widely successful deep learning (DL) has motivated much advancement in HIC \cite{pan2017r,yue2015spectral,zhong2017spectral,chen2014deep,yang2018hyperspectral,mou2017deep}.
DL models like convolution neural networks (CNNs)  acquire parameters via automated training, and are able to extract more representative data features including spatial features.
Though superior to traditional methods in feature extraction and performance, most DL models require large-scale well-annotated training data, which are extremely challenging, even impossible, to attain in hyperspectral image analysis, since annotating a hyper-spectral image often demands professional field exploration, which is unimaginably costly in terms of manpower and time.

In this work, we adopt graph convolutional networks (GCNs) \cite{kipf2016semi} for tackling the challenging HIC task, which operate on a predefined graph, and aggregate and transform feature information from the neighbors of every graph node.
With GCNs, label information is allowed to flow from labeled nodes to unlabeled nodes, thus relaxing the requirement of large-scale training samples while ensuring the performance. 
Several very recent HIC methods have been developed based on GCNs \cite{qin2018spectral,mou2020nonlocal,wan2019multiscale,Hong_2020}.
To the best of our knowledge, \cite{qin2018spectral} is the first to deploy GCN for addressing HIC, where the graph is built by calculating the similarities between each pixel and its adjacent pixels to get spatial context information, and then fed into GCN for computing its label. 
The work~\cite{mou2020nonlocal} argues that the classification of a pixel probably benefits from remote pixels besides its neighbors, and builds a nonlocal graph by measuring similarities of every two hyperspectral pixels in the input HSI.
However, due to the great spectral diversity, such pixel-level graphs cannot precisely reflect the intrinsic relations among the pixels.
To address this issue, \cite{wan2019multiscale} proposes a superpixel-based multi-scale dynamic GCN, which applies superpixel segmentation and multi-scale graph convolution to extensively exploit the spatial information and acquire better feature representation.
Furthermore, in \cite{Hong_2020}, a novel version of GCN called miniGCNs is proposed, where regular patches of the original HSI are used for training the GCN model, yielding lower computation cost.

To further lift classification performance, in this work we design a GCN based HIC framework which applies two rounds of clustering in graph construction to make full use of limited training samples and effectively reduce computation complexity.
In particular, given an input hyperspectral image, we first cluster the pixels with homogeneous spectral features into a superpixel using the hyper manifold simple linear iterative  clustering algorithm (HMS) \cite{sellars2020superpixel}.
This superpixel segmentation operation is also applied in \cite{wan2019multiscale}, which however uses the simple linear iterative clustering (SLIC) \cite{achanta2012slic} to generate superpixels and builds multi-scale GCN branches to capture spatial information of different scales.
Superpixel segmentation is able to directly reduce the size of the initial graph and feature complexity, and also help balance the labeled and unlabeled samples distributions.
We then build a superpixel based graph taking into account the ranking of Euclidean distances of each two nodes, as well as the multi-scale adjacent relationships.
After graph construction, we propose to partition this superpixel level graph into several sub-graphs using the graph partition approach in \cite{karypis1998fast}.
In this process, the unstable edge weights that correspond to relatively weak node dependencies are removed.
This second round of clustering can help our model achieve faster training speed and also efficient convergence.
More importantly, the GCN is enabled to exhaustively exploit the image features and flexibly preserve the class boundaries. 
To capture the long-range dependencies and describe the adjacent intrinsic similarities between nodes more accurately, we focus on the top-k similarity relationships between adjacent multi-hop nodes rather than one hop, and fix the weight edge as 1 to deal with spectral diversity.

We conduct extensive experiments on three popular HSI datasets, and the results well show that our method can achieve the state-of-the-art performance when compared with existing methods.    
To sum up, our main contributions are third-fold.
First, we propose a method to construct a more robust graph based on multi-scale and ranked edge weights to help the GCN model to find reliable dependence.
Second, we design a dual-clustering in graph construction to help our model achieve faster training speed and also efficient convergence.
Third, sub-graphs of HSI are found helpful for HSI classification when only limited training samples are provided.

\begin{figure*}[th]\vspace{0pt}
	\subfigure{\label{fig:frame_work}
		\includegraphics[width=18.0cm]{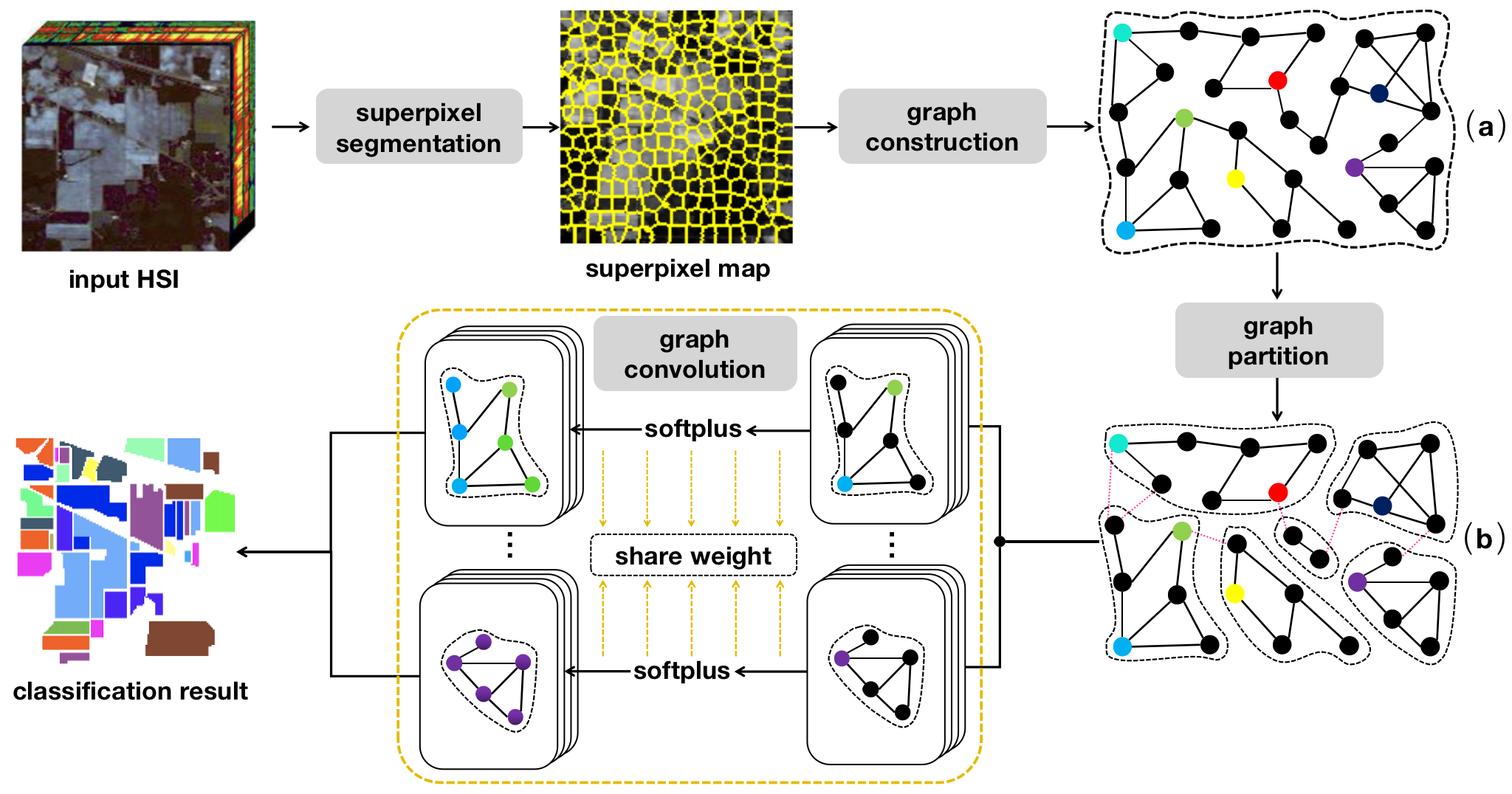}} 
\caption{
Workflow of our method. (a) shows the sketch map for the superpixel level graph, where colorful nodes mean the training samples while  black nodes mean unlabeled nodes. (b) shows the sketch map for the superpixel level sub-graphs, where red lines indicate edges that have been cut off during graph partition.
}
\label{overall workflow}
\end{figure*}

\section{related work}
\subsection{Hyperspectral image classification}

Hyperspectral image classification (HIC) has been studied from long ago in remote sensing, with much prior literature like Bayesian methods \cite{haut2018active}, random forest \cite{ho1998random}, and kernel methods \cite{maji2008classification}.
SVM is shown effective in case of limited labeled examples \cite{mercier2003support} but ignores the correlations among different image pixels. 
Some works address this issue by exploiting spatial information.
To name a few, \cite{camps2006composite} uses SVM with composite kernels for tackling HIC; 
\cite{he2016three} adopts 3D filtering with a Gaussian kernel and its derivative for spectral–spatial information extraction; 
in \cite{he2016discriminative} a discriminative low-rank Gabor filtering method is developed to extract spectral–spatial information.
MRF is widely used to exploit spatial context information by assuming spatially neighboring pixels are more likely to take the same label \cite{li2011spectral}.
But it may suffer performance drop when the neighboring pixels are highly correlated \cite{fauvel2012advances}, and thus conditional random field is applied to directly model the class posterior probability which achieves improved performance \cite{zhang2012simplified}, \cite{zhong2014support}. 
These prior methods adopt hand-crafted spectral–spatial features with heavy dependence on human expertise.

Recently, deep learning based methods \cite{ma2015hyperspectral}, \cite{paoletti2018deep}, \cite{song2018hyperspectral} are successfully applied to HIC, starting from a pioneer work \cite{chen2014deep} which uses stacked autoencoder for extracting high-level features.
Some following works use restricted Boltzmann machine and deep belief network for hyperspectral image feature extraction and pixel classification \cite{li2014classification}, or adopt the RNN model \cite{shi2018multi} to learn the spatial dependence of nonadjacent image patches in a 2-D spatial domain.
CNN based models like \cite{hu2015deep}, \cite{ghamisi2016self} are widely used. 
\cite{yang2016hyperspectral} develops a two-channel deep CNN to jointly learn spectral–spatial features, each learning spectral and spatial features, respectively. 
A deep CNN in \cite{zhang2018diverse} is built to exploit different local or global regions inputs for joint representation learning.
\cite{he2019heterogeneous} incorporates transfer learning to benefit HIC through transferring knowledge of normal pictures to HIS.
Another work \cite{paoletti2020flop} presents a lightweight CNN for tackling HIC, which dramatically reduces the parameters as well as the computational complexity for floating-point operations.
Though effective, these CNN based models simply apply fixed convolution kernels to different regions in a hyperspectral image and ignore the geometric appearance information of various local regions, often leading to misclassification.

\subsection{Graph convolutional networks}

GCN \cite{kipf2016semi} directly operates on a graph and extracts high-level features by aggregating information from neighborhoods of graph nodes.   
It applies to data with arbitrary non-Euclidean structure and has been widely applied \cite{ying2018graph}, \cite{zhang2019star}.

Graph-based HIC methods have been developed starting from \cite{qin2018spectral} that builds the graph on adjacent pixel-level.
However, in \cite{qin2018spectral} the edge weight of the graph is easily affected by the pixel spectral diversity, which cannot accurately reflect the intrinsic relationships of pixels.
The work \cite{wan2019multiscale} uses a superpixel generation technique named SLIC to reduce the number of graph nodes, and builds a GCN that dynamically updates the graph and fuses multi-scale spectral-spatial information to reduce the impact of a bad predefined graph.
However, the conventional texture-based methods exploit contextual relations that are often restricted in a small local region.
In \cite{wan2020hyperspectral}, instead of using superpixel generation,  an adaptively learnable graph projection method is proposed to achieve pixel-to-region assignment and a dynamic graph refinement method is used with a trainable weight matrix to learn an improved distance metric.
Furthermore, \cite{cai2020graph} proposes a graph convolutional subspace clustering (GCSC) framework, taking as input the feature matrix and adjacency matrix of the HSI graph to get a robust graph embedding, then using it as a dictionary for the subsequent affinity learning, and finally outputting classification results by a clustering model.
In another work \cite{mou2020nonlocal}, a nonlocal, data-driven graph representation is proposed, which takes the whole hyperspectral image as input and performs information propagation by a couple of graph convolutions based on a learned nonlocal graph.

Though effective, a dynamic multi-scale GCN architecture contains multi-branches with lots of network parameters and heavy optimization burden. 
In this work, we propose a new HIC framework that adds no network branch for higher performance.
Instead, we incorporate mutli-scale adjacent top-k similarity information to optimize the edge weight of the defined graph, and adopt a graph cluster algorithm (METIS) \cite{karypis1998fast} to remove some unstable edge weights, so as to attain accurate node embedding and thus good classification performance.

\section{proposed method}

Given a label set $ \tau = \left\lbrace1, ..., c \right\rbrace $, 
a set of known observations $\left\lbrace \left(x_{i}, y_{i} \right)  \right\rbrace_{i=1}^{n}$ where $\left\lbrace y_{i} \right\rbrace_{i=1}^{n} \in \tau$, 
and a set of unknown observations $\left\lbrace x_{i} \right\rbrace_{i=n+1}^{n+m}$,
the goal of semi-supervised hyperspectral image classification (HIC) is to find the mapping $ f : \chi\Rightarrow\tau $. 
The $ f $ is expected to give a good prediction of labels $\left\lbrace y_{i} \right\rbrace_{i=1}^{n+m}$ over the union set $\chi\Rightarrow\left\lbrace x_{i} \right\rbrace_{i=1}^{n+m}$ of the labeled and unlabeled data.

In this work, we propose a novel dual-clustering Graph Convolutional Network (GCN) framework for tackling Hyperspectral Image Classification (HIC).
An illustration of its overall workflow is shown in Fig.~\ref{overall workflow}.
Given an input HSI, we first cluster the homogeneous pixels in terms of spectral features into a superpixel, thus forming a set of superpixels.
We then build a superpixel level graph, rather than a pixel level one, to reduce the scale of the initial graph and also the features complexity. 
To this end, we calculate the spectral similarity between the node and multi hop nodes, and select the k neighbor nodes with highest similarities to construct edges in between, so as to better exploit the feature dependency of adjacent nodes.
After that, instead of performing convolution over the whole superpixel level graph for classification, we partition it into several sub-graphs using the approach adopted in a prior method  cluster-GCN~\cite{clustergcn}.
This is based on the consideration that some edges in the initial graph may be accurate.
We hence remove these weak edge weights, and the remained nodes and edges form a set of super-pixel level sub-graphs.
We use these sub-graphs to train the model instead of using the whole.
In this way, the dependence between nodes with high similarity can be strengthened, enabling the model to converge more quickly and accurately.
Also, the scale of the graph is minimized for a second time.
At below, we will elaborate each operation in details.

\subsection{Superpixel segmentation}
Given an input HSI, we first cluster it into a set of meaningful regions (i.e., superpixels), inside which the pixels are similar in terms of spectral features.
Formally, we denote the given HSI as $ \mathbf{I} =\left\lbrace x_{k}\right\rbrace_{k=1}^{w \times h} $ with $\mathit{W \times H \times B} $ dimensions, in which $W$, $H$, $B$ represent the width, height, and the number of bands respectively.
To segment the HSI $ \mathbf{I} $,  the HMS algorithm~\cite{sellars2020superpixel}, which is an extension of the SLIC algorithm~\cite{achanta2012slic}, is adopted to produce accurate superpixels.
Comparing with other superpixel segmentation algorithms, the HMS is more sensitive to the content in HSIs, and the size of its produced superpixels is adjustable by the information density. 
Using the HMS $ \Theta:I \Rightarrow S $, the HSI $\mathbf{I} $ is segmented into a set of superpixels denoted as $ S = \left\lbrace s_{1}^{n_{1}}, s_{2}^{n_{2}}, ..., s_{p}^{n_{p}}\right\rbrace  $ where $ p $ is the number of superpixels, each superpixel $s_{k}^{n_{k}}$ consists of homogeneous pixels $\left\lbrace x_{1}^{k}, x_{2}^{k}, ..., x_{n_{k}}^{k} \right\rbrace$,
and $ \left\lbrace n_{1}, n_{2}, ..., n_{p}\right\rbrace  $ is the number of pixels that are contained in the superpixels, satisfying $\sum_{k=1}^{p} n_{k} = W \times H$.

Above super-pixel segmentation in our method brings several benefits.
First, this operation actually takes advantage of the spatial similarity of adjacent hyperspectral pixels.
By expressing the HSI in the form of superpixels, the spectral diversity of the image can be effectively reduced, which is conducive to classification.
Also, the pixels falling into the same superpixel can share a same label with each other.
That is, if a superpixel contains a labeled sample, the rest pixels in it can be assigned with the same class as the labeled one.
It is very helpful for balancing the positive and negative samples when only limited labeled samples are provided.
Besides, applying superpixel segmentation can significantly reduce the number of nodes in a graph, and thus reduce the computational complexity.

\subsection{Graph construction}
\label{graph construction}
We then transform the HSI into a graph denoted as $ G = \left( \mathbf{V, E}\right) $, which consists of $ N = |V| $ vertices, equal to the number of superpixels, as well as $|E|$ edges with each edge between any two superpixels $i$ and $j$ representing their similarity. 
The notation $A$ is an $N \times N$ sparse matrix with ${i,j}$ denoting the adjacency matrix of $G$,  indicating whether each pair of nodes is connected.
Also, each node (superpixel) is associated with an $ F $-dimensional attribute and  $ X \in \mathbb{R}^{N \times F} $ denotes the attribute matrix for all N nodes.

By exploiting the correlations of neighboring hyperspectral bands, we apply principal component analysis (PCA) to the input HSI to extract non-correlated features with good computational efficiency.
Then we extract the spectral features of the superpixels $ S = \left\lbrace s_{1}^{n_{1}}, s_{2}^{n_{2}}, ..., s_{p}^{n_{p}}\right\rbrace  $ by applying a mean filter to each superpixel to generate the node attribute $ \overrightarrow{S}_{i} $.
For each vector $ \overrightarrow{S}_{i} $, we have \begin{equation}
\overrightarrow{S}_{i} = \frac{\sum_{j=1}^{n_{i}}x_{j}^{i}}{n_{i}}.
\end{equation}

The element $A_{i, j}$ in the adjacency matrix $A$ is usually calculated by a kernel function, such as in MDGCN \cite{wan2019multiscale}:
\begin{equation}
A_{i, j}=\left\{
\begin{array}{cl}
e^{-\gamma\left \|s_i - s_j \right \|^2} & \mathrm{if} \ i \in \mathrm{Nei}(j) \ \mathrm{or} \ j \in \mathrm{Nei}(i)\\
0  &  \mathrm{otherwise} \\
\end{array} \right.
\end{equation} 
where $\gamma$ denotes the continuity tuning parameter, for which it is hard to find an optimal value. 
Though effective for representing the similarity between adjacent nodes, in the above calculation, 
when the spectral features of the nodes are diverse, the spectral distance between them is unstable, which is prone to cause a bad predefined graph.
To tackle this problem, we propose to select the top-k spectral similar nodes in the surrounding nodes, and keep the weight $A_{i, j}$ as 1.
In this way, we make the current node depend on the adjacent nodes with strong correlations, and the discrete k value is more conducive to finding the optimal value. 
We measure $A_{i, j}$ by
\begin{equation}
A^{k}_{i, j}=\left\{
\begin{array}{cl}
1 & \mathrm{if} \ i \in \mathrm{Near(Nei}(j), k) \ \mathrm{or} \ \mathrm{inverse}\\
0  &  \mathrm{otherwise.} \\
\end{array} \right.\
\label{func1}
\end{equation} 
And the distance between each two superpixels is the Euclidean speactral distance:
\begin{equation}
dis_{i, j} = \|\overrightarrow{S}_{i} - \overrightarrow{S}_{j}\|_{2}.
\end{equation}

\begin{figure}[t]

\begin{minipage}[b]{1.0\linewidth}
  \centering
 \centerline{\epsfig{figure=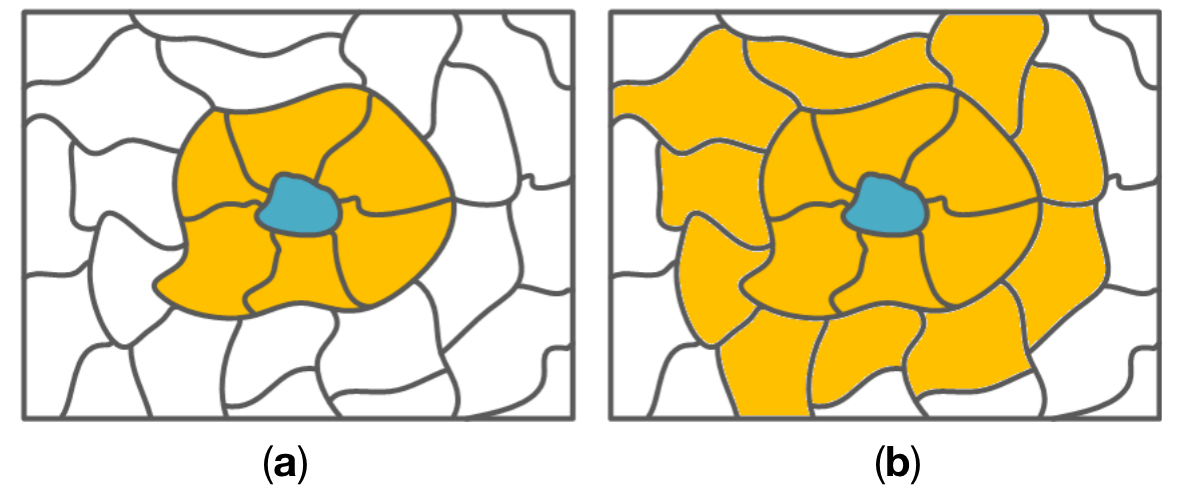,width=8cm}}
\end{minipage}
\caption{Illustration of multi-scales considered by our method. Best viewed in color. Irregular blocks represent superpixels. (a) (b) represent one, two order of neighbors respectively used in our method.}
\label{multi-scale adjacent}

\end{figure}

According to previous works \cite{7729625, wan2019multiscale}, multiscale information is useful for hyperspectral image classification problems. 
This is because the objects in a hyperspectral image usually present different geometric appearances, and the contextual information revealed by different scales is helpful to the model in exploiting the abundant local property of the image regions from diverse levels. 
In our method, we also adopt a multi-order adjacent matrix, as shown in Fig. \ref{multi-scale adjacent}, to define a more robust graph.
Combining with the multi-scale neighbours, we extend $A^{k}_{i, j}$ into $A^{h,k}$:
\begin{equation}
\label{eq1}
A^{h, k}_{i, j}=\left\{
\begin{array}{cl}
1 &  i \in \mathrm{Near(Nei}(j, h), k)\ \mathrm{or}\ \mathrm{inverse}\\
0  &  \mathrm{otherwise} \\
\end{array} \right.
\end{equation} 
where Nei($j, h$) represent the neighbour nodes within $h$ hops.

To combine the weights of different hop adjacent matrices $A^{h,k}$ and highlight the nodes within the top-k range in different scales, we apply an element-wise sum to get more robust dependencies, as blow:
\begin{equation}
\label{eq2}
A_{o}^{k} = \sum_{i=1}^{o}A^{i, k}.
\end{equation}
where $o$ denotes the scale of the order of nodes covered in our adjacency matrix.

\subsection{Graph partition}
\label{graph partition}
Till here, we have obtained a graph $G$ from the input HSI.
However, the edges in the graph $G$ we have constructed inevitably contain error due to the complexity of the actual spectrum.
To reduce the impact of such error as much as possible, we try to prune the weak edges in the graph.

In a prior method cluster-GCN \cite{clustergcn}, a graph cluster algorithm (METIS)~\cite{karypis1998fast} is used to partition the nodes of the graph into several batches for achieving efficient memory usage on large-scale graphs, especially on deep GCNs.
We are then inspired to apply METIS to partition our superpiexel level graph $G$.
In METIS, the number of edges whose incident vertices belong to different partitions is called the edge-cut of the partition. 
METIS can produce partitions with substantially smaller edge-cut, which means the unimportant edges would be reduced during the process of clustering.

The METIS algorithm is very suitable for segmenting our superpixel level graph. 
This is because the edges in our superpixel graph are constructed according to the spatial adjacency, and the sub-graphs generated by METIS are also continuous in space.
Thus applying METIS to our graph produces little impact on the spatial information of the latter. 
Also, in the process of graph partition, METIS will produce partition sets with the least cost, and the edges with lower cost have smaller weight in the graph. 
Cutting off these smaller weight edges is conducive to accelerating the convergence of the model and increasing the classification ability of the model.

Formally, we use METIS to partition the superpixel level graph $G$ to get $ c $ clusters $ \mathcal{V}_{1}, \mathcal{V}_{2},  \mathcal{V}_{3},  ..., \mathcal{V}_{c} $ based on the proposed adjacent matrix $ A $, where $\mathcal{V}_{i}$ represents the vertices subset after clustering, and satisfies $\sum_{i=1}^{c} |\mathcal{V}_{i}| = N$. 
Subsequently, we use the vertices in the clusters $ \mathcal{V}_{1}, \mathcal{V}_{2},  \mathcal{V}_{3},  ..., \mathcal{V}_{c} $ to build the sub-graph set ${G_{1}=(V_{1}, E_{1}), G_{2}=(V_{2}, E_{2}), G_{3}=(V_{3}, E_{3}), ..., G_{c}=(V_{c}, E_{c})}$, and their corresponding adjacent matrix ${A_{1}, A_{2}, A_{3} ..., A_{c}}$ and corresponding attribute matrix $X_{1}, X_{2}, X_{3}, ..., X_{c}$.

\subsection{Graph classification}

We then conduct classification based on the obtained sub-graphs.
GCN is a multilayer neural network, which operates directly on a graph and generates node embeddings by gradually fusing the features in the neighborhood.
Different from a traditional CNN that only applies to the data represented by regular grids, GCN is applicable to  the data with arbitrary non-Euclidean structure.
During training, we randomly select a sub-graph from the sub-graphs set as the input of the GCN model. 
In particular, taking in a sub-graph $G_{i}=(V_{i}, E_{i})$, and the corresponding attribute matrix $X_{i}$ and adjacency matrix $A_{i}$, an L-layer GCN consists of L graph convolution layers and each of them constructs node embedding for the next layer depending on the embedding of the node’s neighbors in the graph $G$ from the previous layer: 
\begin{equation}
 Z^{(l+1)}_{i} = A^{'}_{i}X^{(l)}_{i}W^{(l)} 
\end{equation}
\begin{equation}
X^{(l+1)}_{i} = \theta(Z^{(l+1)}_{i})
\end{equation}
where $ X^{(l)}_{i} \in \mathbb{R}^{|V_{i}| \times F_{l}}$ is the embedding at the $ l $-th layer for all the $ |V_{i}| $ nodes in the sub-graph $G_{i}$, and $ X^{0}_{i} = X_{i} $; $ A^{'}_{i} $ is the normalized and regularized adjacency matrix and $ W^{(l)} \in \mathbb{R}^{F_{l} \times F_{l+1}} $ is the feature transformation matrix which will be learnt during the gradient back-propagation and it is shared between different sub-graphs;
$ \theta(\cdot) $ means the activation function behind each layer, and in this work we adopt the RELU.
The normalized adjacency matrix $ A^{'}_{i} $ is calculated by
\begin{equation}
\check{A_{i}} = A_{i} + I
\end{equation}
\begin{equation}
A^{'}_{i} = D^{-\frac{1}{2}}\check{A_{i}}D^{-\frac{1}{2}}
\end{equation}
where $ D $ is the degree matrix of $ \check{A_{i}} $.

In actual deployment of GCN, we add a 1$\times$1 convolution layer as the first layer of the backbone to change the dimension of the node attribution and also enhance the classification capacity of the model. 
Since deep GCNs may not converge, we fix the number of graph convolution layers as two. 
In every training step, we get the nodes prediction results $ Z_{c} $ of the sub-graph $G_{c}=(A_{c}, E_{c})$:
\begin{equation}
Z_{c} = \mathrm{softmax}(\check{A}_{c} \mathrm{RELU}(\check{A}\Theta(X_{c}W^{(0)}))W^{(1)})
\label{graph_conv}
\end{equation}
where $ \Theta $ represents the $ 1 \times 1 $ convolution operator, and $X_{c}$ represents the corresponding attribute matrix. 
In our model, the cross-entropy loss $\boldsymbol{\Gamma}$ is adopted to penalize the difference between the network output and labels of the original labeled samples, which is \begin{equation}
\boldsymbol{\Gamma} = - \sum_{g\in \mathbf{y}_{G}}\sum_{f=1}^{\tau}\mathbf{Y}_{gf}\mathrm{ln} \mathbf{O}_{gf}
\label{loss_func}
\end{equation} where $ \mathbf{Y}_G $ is the set of indices corresponding to labeled examples, $ \tau $ denotes the label set, and $ \mathbf{Y} $ denotes the label matrix. 
During the test, we input all the sub-graphs to get the classification results of all superpixels, which are then mapped back to the original image to get the final classification result.
The implementation details of our framework are shown in Algorithm \ref{proposed framework algorithm}.

\begin{algorithm}[h]
\caption{Proposed framework}
\label{proposed framework algorithm}
\LinesNumbered
\KwIn{Input image $ I $, label $ Y $}
\KwOut{Predict label $ \check{Y} $ for each pixel in $ I $}
Segment $ I $ into superpixels set $ S $ via HMS algorithm\; 
Construct initial graph $ G=(V, E) $ and adjacent matrix $A$, by using the proposal method that described in Section B\;
Partition graph nodes $ V $ into $ c $ clusters $ \mathcal{V}_{1},  \mathcal{V}_{2}, ..., \mathcal{V}_{c}$ by METIS, then form sub-graph set $\textbf{G}=(G_{1}, G_{2}, G_{3}, ..., G_{c})$\;
// Train GCN model\;
\For{$ iter = 1, ..., max\_iter $}{
Randomly choose one sub-graph $G_{i}$ from $\textbf{G}$\;
Conduct graph convolution according to \ref{graph_conv}\;
Compute loss on the sub-graph $G_{i}$ according to \ref{loss_func}\;
Conduct Adam optimization\;
}
Calculate the network output $ O $ and labels of pixel $ \tau $
\end{algorithm}

\section{experiments}
We conduct extensive experiments to validate the effectiveness of our proposed framework.

\subsection{Experiment settings}

\begin{table}[h] 
\centering
 \caption{\label{tab:indian_pines}INDIAN PINES DATASET SAMPLES STATISTICS} 
 \begin{tabular}{ccccccc} 
  \toprule
Label ID& Class& Labeled& Unlabeled \\
  \midrule 
1& Alfalfa & 30 & 16 \\
2& Corn-notill & 30 & 1398 \\
3& Corn-mintill & 30 & 800 \\
4& Corn & 30 & 207 \\
5& Grass-pasture & 30 & 453 \\
6& Grass-trees & 30 & 700 \\
7& Grass-pasture-mowed & 15 & 13 \\
8& Hay-windrowed & 30 & 448 \\
9& Oats & 15 & 5 \\
10& Soybean-notill & 30 & 942 \\
11& Soybean-mintill & 30 & 2425 \\
12& Soybean-clean & 30 & 563 \\
13& Wheat & 30 & 175 \\
14& Woods & 30 & 1235 \\
15& Buildings-grass-trees-drives & 30 & 356 \\
16& Stone-steel-towers & 30 & 63 \\
  \bottomrule 
 \end{tabular} 
\end{table}
We adopt three popular benchmark datasets in the experiments, which are introduced as follows.

\begin{figure}[h]

\begin{minipage}[b]{1.0\linewidth}
  \centering
 \centerline{\epsfig{figure=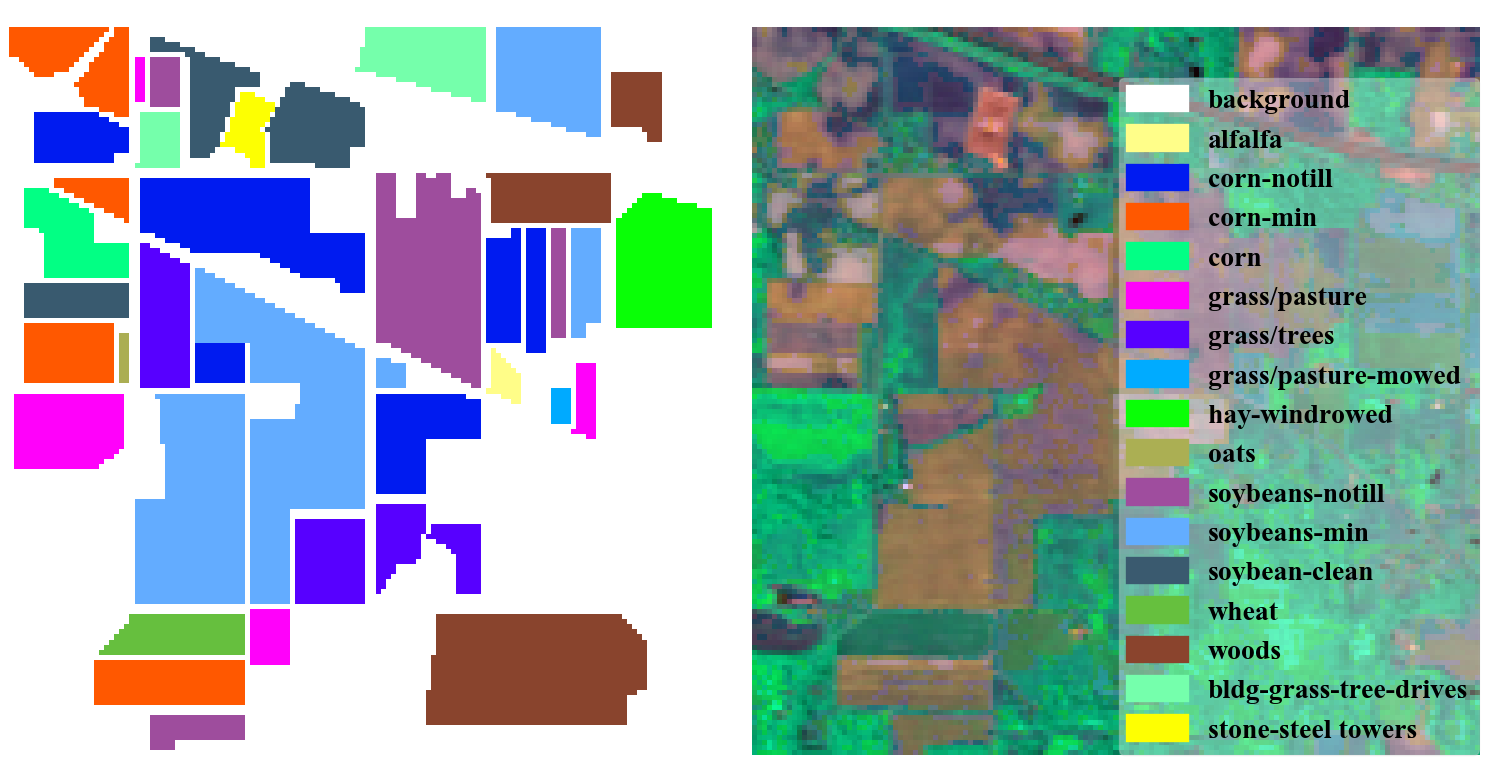,width=8.8cm}}
\end{minipage}
\caption{Indian Pines. Left: Groud-truth map. Right: False-color image.}
\label{Indian Pines dataset}

\end{figure}

$\bullet$ $ \textit{Indian Pines}$ (abbreviated as IP): 
This dataset was acquired by Airborne Visible/Infrared Imaging Spectrometer (AVIRIS) sensor over Indian Pines test site in north-western Indiana in 1992 and has 16 classes. 
Fig. \ref{Indian Pines dataset} shows the false-color image and its ground truth.
Origin data consists of 145$\times$145 pixels, 220 spectral bands ranging from 0.4 to 2.5$\mu$m, with a spatial resolution of 20m. 
After removing 20 bands covering the region of water absorption, we use the remaining 200 bands data for our experiments.

\begin{table}[h] 
\centering
 \caption{\label{tab:pavia}UNIVERSITY OF PAVIA DATASET SAMPLES NUMBERS} 
 \begin{tabular}{ccccccc} 
  \toprule
Label ID& Class& Labeled& Unlabeled \\
  \midrule 
1& Asphalt & 30 & 6601 \\
2& Meadows & 30 & 18619 \\
3& Gravel & 30 & 2069 \\
4& Trees & 30 & 3034 \\
5& Painted metal sheets & 30 & 1315 \\
6& Bare soil & 30 & 4999 \\
7& Bitumen & 30 & 1300 \\
8& Self-blocking bricks & 30 & 3652 \\
9& Shadows & 30 & 917 \\
  \bottomrule 
 \end{tabular} 
 \label{indian samples}
\end{table}

\begin{table}[!h] 
\centering
 \caption{\label{tab:pavia}SALINAS DATASET SAMPLES NUMBERS} 
 \begin{tabular}{ccccccc} 
  \toprule
Label ID& Class& Labeled& Unlabeled \\
  \midrule 
1& Brocoli green weed1 & 30 & 1979 \\
2& Brocoli green weed 22 & 30 & 3696 \\
3& Fallow & 30 & 1946 \\
4& Fallow rough plow & 30 & 1364 \\
5& Fallow smooth & 30 & 2648 \\
6& Stubble & 30 & 3929 \\
7& Celery & 30 & 3549 \\
8& Grapes untrained & 30 & 11241 \\
9& Soil vinyard develop & 30 & 6173 \\
10& Corn senesced greeen weeds & 30 & 3248 \\
11& lettuce romaine 4wk & 30 & 1038 \\
12& lettuce romaine 5wk & 30 & 1897 \\
13& lettuce romaine 6wk & 30 & 886 \\
14& lettuce romaine 7wk & 30 & 1040 \\
15& Vinyard untrained & 30 & 7238 \\
16& Vinyard vertical trellis & 30 & 1777 \\
  \bottomrule 
 \end{tabular} 
 \label{salinas samples}
\end{table}

\begin{figure}[h]

\begin{minipage}[b]{1.0\linewidth}
  \centering
 \centerline{\epsfig{figure=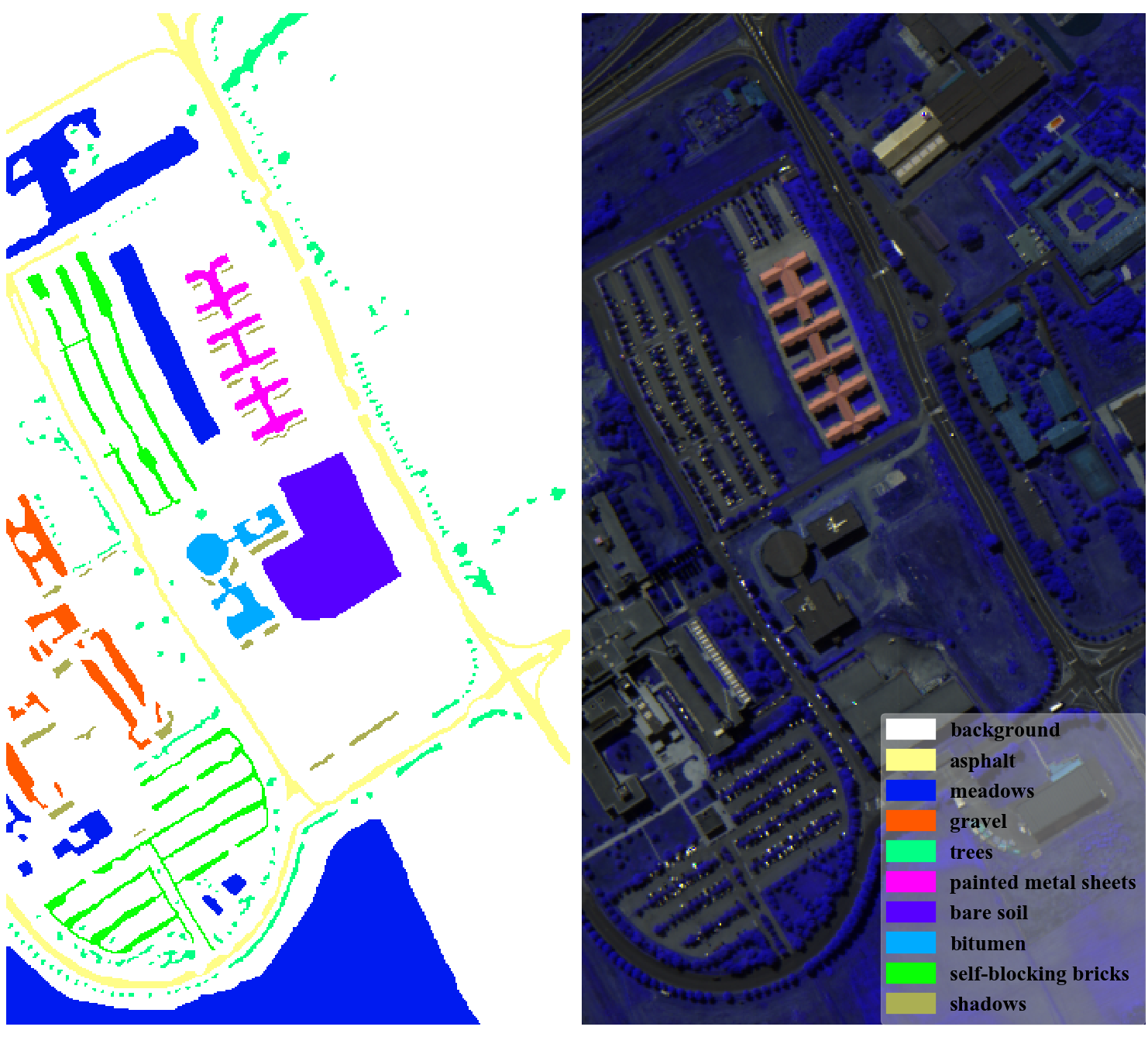,width=7.6cm}}
\end{minipage}
\caption{University of Pavia. Left: Groud-truth map. Right: False-color image.}
\label{pavia dataset}

\end{figure}

\begin{figure}[!h]

\begin{minipage}[b]{1.0\linewidth}
  \centering
 \centerline{\epsfig{figure=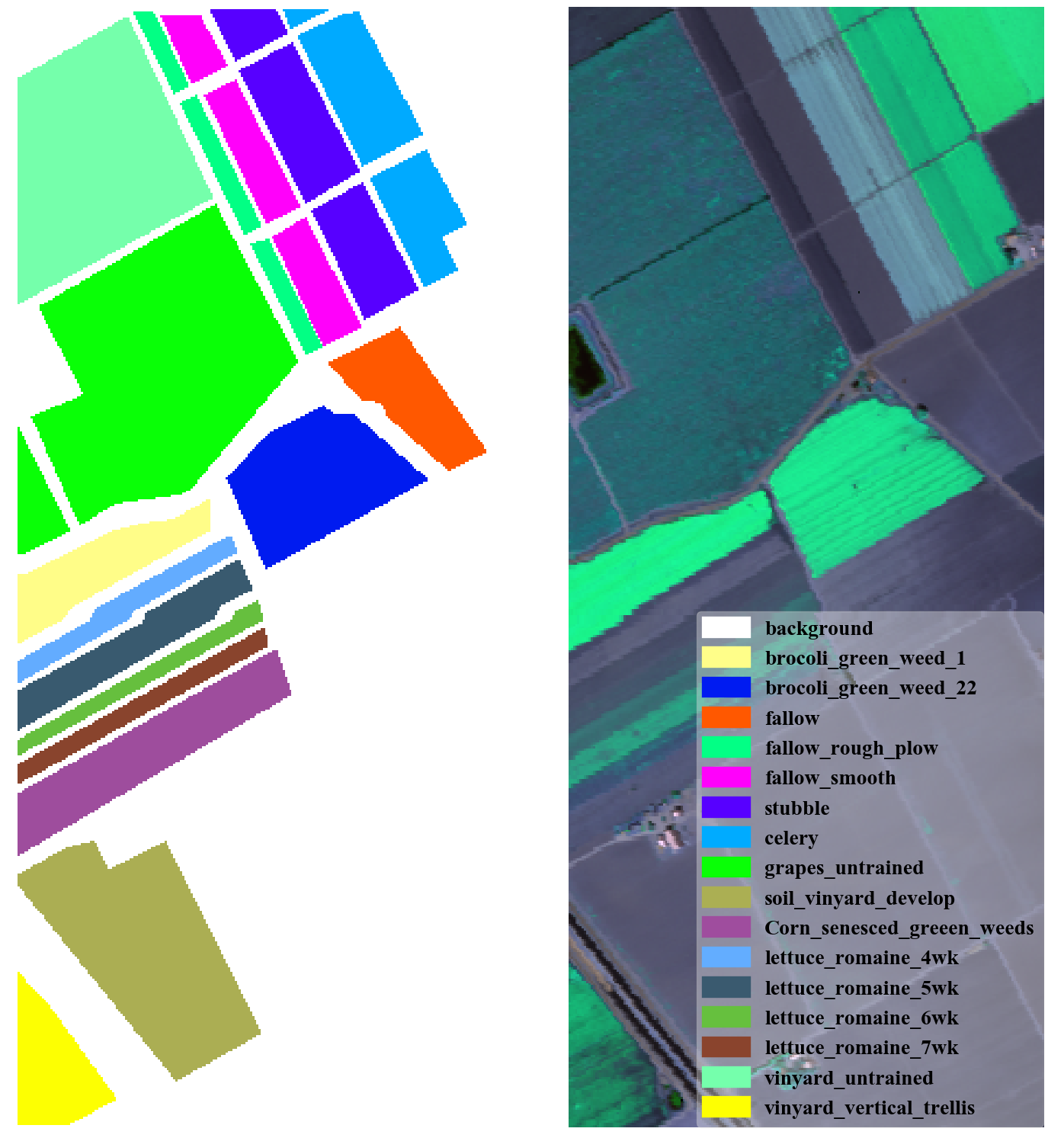,width=7.8cm}}
\end{minipage}
\caption{Salinas. Left: Groud-truth map. Right: False-color image.}
\label{salinas dataset}

\end{figure}

$\bullet$ $\textit{University of Pavia}$ (abbreviated as UP): The dataset was collected by the Reflective Optics System Imaging Spectrometer (ROSIS) over the Pavia University in Italy in 2001 and contains 9 classes as shown as Fig. \ref{pavia dataset}.
It consists of 610$\times$340 pixels with a spatial resolution of 1.3m, 103 spectral bands in the wavelength range from 0.43 to 0.86$\mu$m.

$\bullet$ $\textit{Salinas}$ (abbreviated as SA): 
This image was taken by AVIRIS sensor over Salinas Valley, California, and contains 16 classes.
Fig. \ref{salinas dataset} shows the false-color image and its ground truth. 
The image size is 512$\times$217 pixels with a spatial resolution of 3.7m and has 224 spectral channels over 0.4 to 2.5$\mu$m.
After a usual water absorption region removing, we use the remaining 200 channels as experimental data.

\begin{table*}[h]
\renewcommand\arraystretch{1.3}
\centering
\caption{CLASSIFICATION RESULTS OF DIFFERENT APPROACHES ON INDIAN PINE DATASET}\label{tab:aStrangeTable}
\vspace{0.2cm}
\begin{tabular}{ccccccccc}
\toprule
ID& EMAP\cite{plaza2004new}& pResNet\cite{paoletti2018deep}& MDGCN \cite{wan2019multiscale}& S$^{2}$GCN\cite{qin2018spectral}& FDSSC\cite{wang2018fast}& RNN\cite{mou2017deep}& SSRN\cite{zhong2017spectral}& Proposed\\
\midrule
1& 97.82$\pm$0.97& \textbf{100.00$\pm$0.00}& 98.75$\pm$2.50& \textbf{100.00$\pm$0.00}& 81.20$\pm$17.83& 90.00$\pm$9.35& 93.24$\pm$2.63& \textbf{100.00$\pm$0.00}\\
2& 77.06$\pm$4.67& 77.34$\pm$3.38& 84.48$\pm$3.42& 84.43$\pm$2.50& 86.52$\pm$8.28& 51.83$\pm$3.75& 76.63$\pm$5.96& \textbf{91.28$\pm$3.60}\\
3& 85.43$\pm$3.12& 87.83$\pm$5.68& 87.36$\pm$5.58& 82.87$\pm$5.53& 87.59$\pm$8.14& 46.35$\pm$4.64& 68.78$\pm$7.53& \textbf{92.88$\pm$3.96}\\
4& 95.10$\pm$2.60& \textbf{98.81$\pm$1.98}& 97.29$\pm$2.84& 93.08$\pm$1.95& 72.13$\pm$13.50& 68.98$\pm$9.14& 87.64$\pm$2.49& 98.11$\pm$1.51\\
5& 86.35$\pm$3.56& 93.80$\pm$2.96& 93.66$\pm$3.02& 97.13$\pm$1.34& \textbf{98.94$\pm$1.09}& 85.03$\pm$3.79& 86.72$\pm$1.54& 95.54$\pm$3.39\\
6& 91.34$\pm$3.04& 96.42$\pm$2.71& 97.74$\pm$2.28& 97.29$\pm$1.27& 98.52$\pm$0.89& 91.35$\pm$2.92& 92.05$\pm$1.82& \textbf{98.67$\pm$1.04}\\
7& 95.71$\pm$3.49& \textbf{100.00$\pm$0.00}& \textbf{100.00$\pm$0.00}& 92.31$\pm$0.00& 64.39$\pm$29.55& 92.30$\pm$6.88& 95.66$\pm$0.51& \textbf{100.00$\pm$0.00}\\
8& 98.20$\pm$1.30& 99.40$\pm$0.85& \textbf{100.00$\pm$0.00}& 99.03$\pm$0.93& 99.95$\pm$0.08& 96.29$\pm$1.33& 95.90$\pm$2.97& \textbf{100.00$\pm$0.00}\\
9& 99.50$\pm$1.50& \textbf{100.00$\pm$0.00}& \textbf{100.00$\pm$0.00}& \textbf{100.00$\pm$0.00}& 56.33$\pm$25.95& 98.00$\pm$6.00& \textbf{100.00$\pm$0.00}& \textbf{100.00$\pm$0.00}\\
10& 82.53$\pm$4.06& 87.51$\pm$4.59& 88.59$\pm$3.33& \textbf{93.77$\pm$3.73}& 83.91$\pm$5.49& 58.99$\pm$5.92& 82.42$\pm$3.24& 91.91$\pm$3.78\\
11& 75.04$\pm$4.73& 80.19$\pm$2.71& 79.26$\pm$3.47& 84.98$\pm$2.82& \textbf{94.55$\pm$3.44}& 60.09$\pm$6.84& 82.23$\pm$2.88& 91.79$\pm$3.79\\
12& 86.76$\pm$2.18& 79.72$\pm$7.25& \textbf{90.28$\pm$6.48}& 80.05$\pm$5.17& 87.44$\pm$7.23& 63.97$\pm$7.16& 69.09$\pm$4.36& 90.17$\pm$5.54\\
13& 98.68$\pm$0.38& \textbf{99.87$\pm$0.25}& 99.48$\pm$0.17& 99.43$\pm$0.00& 97.99$\pm$2.54& 98.40$\pm$1.16& 95.78$\pm$0.75& 99.65$\pm$0.27\\
14& 89.62$\pm$5.11& 95.57$\pm$2.58& 95.81$\pm$2.99& 96.73$\pm$0.92& 99.02$\pm$0.48& 86.21$\pm$3.84& 86.52$\pm$2.43& \textbf{99.73$\pm$0.66}\\
15& 92.95$\pm$4.64& 95.62$\pm$2.75& 99.46$\pm$0.7& 86.80$\pm$3.42& 84.95$\pm$6.67& 66.68$\pm$5.63& 73.12$\pm$5.28& \textbf{99.94$\pm$0.16}\\
16& 96.23$\pm$2.10& 99.12$\pm$0.87& \textbf{100.00$\pm$0.00}& \textbf{100.00$\pm$0.00}& 82.34$\pm$4.35& 93.01$\pm$3.11& 86.21$\pm$1.30& \textbf{100.00$\pm$0.00}\\
\midrule
OA& 84.13$\pm$1.28& 87.18$\pm$0.60& 88.93$\pm$1.70& 89.49$\pm$1.08& 90.79$\pm$1.93& 67.67$\pm$1.15& 88.34$\pm$1.73& \textbf{94.65$\pm$1.21}\\
AA& 90.52$\pm$0.83& 93.20$\pm$0.50& 94.51$\pm$0.66& 92.99$\pm$1.04& 85.99$\pm$2.48& 77.97$\pm$1.35& 85.75$\pm$0.69& \textbf{96.85$\pm$0.40}\\
Kappa& 83.50$\pm$1.31& 85.41$\pm$0.68& 88.43$\pm$1.56& 88.00$\pm$1.23& 89.52$\pm$2.18& 63.22$\pm$1.20& 86.68$\pm$1.98& \textbf{94.4$\pm$1.46}\\
\bottomrule
\end{tabular}
\label{indian exp tab}
\end{table*}

\begin{table*}[h]
\renewcommand\arraystretch{1.3}
\centering
\vspace{0.2cm}
\caption{CLASSIFICATION RESULTS OF DIFFERENT APPROACHES ON UNIVERSITY OF PAVIA DATASET}\label{tab:aStrangeTable}
\vspace{0.2cm}
\begin{tabular}{ccccccccc}
\toprule
ID& EMAP\cite{plaza2004new}& pResNet\cite{paoletti2018deep}& MDGCN \cite{wan2019multiscale}& S$^{2}$GCN\cite{qin2018spectral}& FDSSC\cite{wang2018fast}& RNN\cite{mou2017deep}& SSRN\cite{zhong2017spectral}& Proposed\\
\midrule
1& 94.35$\pm$1.13& 89.18$\pm$3.25& 86.02$\pm$3.29& 92.78$\pm$3.79& \textbf{99.14$\pm$0.62}& 74.19$\pm$7.61& 98.80$\pm$1.10& 92.85$\pm$3.51\\
2& 75.31$\pm$5.50& 96.80$\pm$2.16& 94.90$\pm$2.09& 87.06$\pm$4.47& \textbf{99.41$\pm$0.38}& 87.96$\pm$1.19& 98.45$\pm$0.54& 97.53$\pm$1.40\\
3& 92.05$\pm$6.64& 86.95$\pm$4.47& 95.29$\pm$3.04& 87.97$\pm$4.77& 85.68$\pm$8.79& 60.27$\pm$10.77& 77.05$\pm$10.24& \textbf{97.94$\pm$1.18}\\
4& \textbf{97.57$\pm$1.94}& 96.47$\pm$1.92& 93.09$\pm$2.34& 90.85$\pm$0.94& 94.38$\pm$5.63& 81.54$\pm$4.58& 83.02$\pm$9.07& 94.57$\pm$1.09\\
5& 96.47$\pm$3.58& 99.61$\pm$0.81& 99.11$\pm$0.89& \textbf{100.00$\pm$0.00}& 99.92$\pm$0.06& 99.62$\pm$0.21& 99.96$\pm$0.09& 99.49$\pm$0.68\\
6& 86.43$\pm$8.73& 88.51$\pm$5.47& \textbf{99.77$\pm$0.18}& 88.69$\pm$2.64& 89.19$\pm$4.97& 76.20$\pm$5.96& 87.03$\pm$6.26& 98.57$\pm$2.78\\
7& 95.35$\pm$0.35& 88.29$\pm$5.05& 97.26$\pm$1.77& 98.88$\pm$1.08& 91.62$\pm$3.44& 76.38$\pm$7.32& 83.92$\pm$8.97& \textbf{100.00$\pm$0.00}\\
8& 86.07$\pm$6.46& 89.34$\pm$5.07& 95.34$\pm$2.04& 89.97$\pm$3.28& 93.74$\pm$1.31& 89.90$\pm$2.64& 88.41$\pm$4.63& \textbf{96.00$\pm$2.77}\\
9& 94.54$\pm$3.05& 99.39$\pm$0.49& 98.69$\pm$1.06& 98.89$\pm$0.53& 99.88$\pm$0.19& \textbf{100.00$\pm$0.00}& 99.97$\pm$0.04& 97.51$\pm$1.40\\
\midrule
OA& 84.69$\pm$1.84& 93.38$\pm$1.12& 94.30$\pm$1.04& 89.74$\pm$1.70& 96.06$\pm$1.29& 83.06$\pm$2.54& 92.81$\pm$1.90& \textbf{96.87$\pm$1.11}\\
AA& 91.13$\pm$1.02& 92.72$\pm$1.41& 95.50$\pm$0.52& 92.80$\pm$0.47& 94.77$\pm$1.70& 82.89$\pm$2.17& 90.73$\pm$2.26& \textbf{97.16$\pm$0.76}\\
Kappa& 84.27$\pm$1.86& 91.23$\pm$1.48& 94.09$\pm$1.06& 86.65$\pm$2.06& 94.82$\pm$1.66& 77.71$\pm$3.09& 90.59$\pm$2.44& \textbf{96.77$\pm$1.27}\\
\bottomrule
\end{tabular}
\label{pavia exp tab}
\end{table*}

For all the three data sets, 30 labeled pixels are randomly selected in all classes for training, and only 15 labeled samples are chosen if the corresponding class has less than 30 examples.
Besides, we use  90$\%$ of labeled samples to adjust classifier parameters during the training, and the rest 10$\%$ labeled samples serve as the validation set. 
All unlabeled samples are used as the test set to evaluate the classification performance.

Several recent state-of-the-art HSI classification methods are used for comparison with our method, including EMAP~\cite{plaza2004new}, pResNet~\cite{paoletti2018deep}, FDSSC~\cite{wang2018fast}, RNN~\cite{mou2017deep}, SSRN~\cite{zhong2017spectral}, MDGCN~\cite{wan2019multiscale}, S$^{2}$GCN~\cite{qin2018spectral}. 
Additionally, in order to evade the effect of superpixel segmentation, we conduct our work and MDGCN on a same super-pixel graph.  
Three evaluation protocols, including overall accuracy (OA), average accuracy (AA), and Kappa Coefficient (K) are used to evaluate the performance of each method.

\begin{table*}[!h]
\renewcommand\arraystretch{1.3}
\centering
\caption{CLASSIFICATION RESULTS OF DIFFERENT APPROACHES ON SALINAS DATASET}\label{tab:aStrangeTable}
\vspace{0.1cm}
\begin{tabular}{ccccccccc}
\toprule
D& EMAP\cite{plaza2004new}& pResNet\cite{paoletti2018deep}& MDGCN \cite{wan2019multiscale}& S$^{2}$GCN\cite{qin2018spectral}& FDSSC\cite{wang2018fast}& RNN\cite{mou2017deep}& SSRN\cite{zhong2017spectral}& Proposed\\
\midrule
1& 99.71$\pm$0.09& 99.62$\pm$0.50& 99.90$\pm$0.08& -& 100.00$\pm$0.00& 97.99$\pm$1.67& 99.73$\pm$0.14& \textbf{99.97$\pm$0.04}\\
2& 98.65$\pm$0.87& 99.84$\pm$0.30& \textbf{100.00$\pm$0.00}& -& \textbf{100.00$\pm$0.00}& 99.08$\pm$1.45& \textbf{100.00$\pm$0.00}& \textbf{100.00$\pm$0.00}\\
3& 99.74$\pm$0.10& 99.69$\pm$0.55& 99.74$\pm$0.12& -& 97.69$\pm$2.30& 96.27$\pm$3.74& \textbf{100.00$\pm$0.00}& 99.92$\pm$0.11\\
4& 99.20$\pm$0.73& 98.86$\pm$1.41& 97.36$\pm$2.11& -& 97.71$\pm$0.71& 99.20$\pm$0.36& \textbf{99.23$\pm$0.21}& 97.99$\pm$3.06\\
5& 96.99$\pm$0.30& 98.26$\pm$1.44& 98.56$\pm$0.32& -& \textbf{99.93$\pm$0.07}& 96.15$\pm$1.74& 98.34$\pm$1.03& 97.66$\pm$1.54\\
6& 97.09$\pm$1.84& 99.98$\pm$0.03& 99.57$\pm$0.30& -& 99.98$\pm$0.02& 99.43$\pm$0.36& 98.62$\pm$0.54& \textbf{100.00$\pm$0.00}\\
7& 98.34$\pm$0.95& 99.67$\pm$0.28& 99.52$\pm$0.25& -& 99.85$\pm$0.36& 99.40$\pm$0.17& 99.77$\pm$0.02& \textbf{100.00$\pm$0.00}\\
8& 74.94$\pm$6.61& 85.33$\pm$3.47& 95.68$\pm$3.13& -& 91.53$\pm$3.69& 80.42$\pm$2.96& 82.49$\pm$7.26& \textbf{97.07$\pm$2.49}\\
9& 98.96$\pm$0.30& 99.61$\pm$0.51& 99.89$\pm$0.04& -& 99.48$\pm$0.18& 98.99$\pm$0.69& 99.24$\pm$0.34& \textbf{99.99$\pm$0.02}\\
10& \textbf{97.44$\pm$2.25}& 96.37$\pm$2.07& 96.49$\pm$1.17& -& 95.77$\pm$2.84& 89.07$\pm$2.63& 95.04$\pm$2.86& 94.97$\pm$1.70\\
11& 96.90$\pm$1.40& 99.37$\pm$0.69& \textbf{100.00$\pm$0.00}& -& 97.51$\pm$2.11& 94.19$\pm$2.14& 99.01$\pm$0.74& 98.96$\pm$1.46\\
12& \textbf{99.95$\pm$0.02}& 99.61$\pm$0.82& 99.68$\pm$0.31& -& 99.83$\pm$0.16& 99.51$\pm$0.51& 99.13$\pm$0.32& 99.15$\pm$1.57\\
13& 98.13$\pm$0.84& \textbf{99.92$\pm$0.16}& 92.78$\pm$5.32& -& 99.68$\pm$0.39& 98.34$\pm$0.61& 98.22$\pm$1.22& 98.60$\pm$0.19\\
14& 93.34$\pm$3.46& \textbf{99.59$\pm$0.78}& 92.69$\pm$2.19& -& 98.97$\pm$1.48& 96.19$\pm$1.76& 99.30$\pm$0.09& 97.38$\pm$1.40\\
15& 77.62$\pm$2.98& 86.69$\pm$4.93& \textbf{99.09$\pm$0.34}& -& 78.98$\pm$5.72& 57.05$\pm$4.74& 87.62$\pm$5.66& 97.88$\pm$1.10\\
16& 97.34$\pm$3.25& 98.28$\pm$1.91& 90.38$\pm$6.72& -& 99.97$\pm$0.07& 97.07$\pm$1.18& 96.37$\pm$1.34& \textbf{100.00$\pm$0.00}\\
\midrule
OA& 90.58$\pm$1.07& 94.60$\pm$0.83& 97.94$\pm$1.13& -& 94.47$\pm$0.96& 88.43$\pm$0.92& 94.65$\pm$1.78& \textbf{98.67$\pm$0.60}\\
AA& 95.27$\pm$0.25& 97.54$\pm$0.83& 97.58$\pm$0.37& -& 97.31$\pm$0.37& 93.65$\pm$0.63& 97.11$\pm$0.35& \textbf{98.72$\pm$0.37}\\
Kappa& 90.23$\pm$1.12& 93.99$\pm$0.92& 97.85$\pm$1.09& -& 93.86$\pm$1.06& 87.10$\pm$1.02& 94.04$\pm$1.66& \textbf{98.38$\pm$0.75}\\
\bottomrule
\end{tabular}
\label{salinas exp tab}
\end{table*}

\begin{figure*}[!htb]\vspace{10pt}
	\centering
	\subfigure[Ground-truth]{\label{fig:results_of_qb_IN}
		\includegraphics[width=3.2cm]{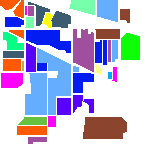}} \hspace{12pt}\vspace{2pt}
	\subfigure[EMAP\cite{plaza2004new}]{\label{fig:results_of_qb_IN}
		\includegraphics[width=3.2cm]{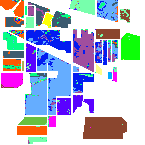}} \hspace{12pt}\vspace{2pt}
	\subfigure[PResNet\cite{paoletti2018deep}]{\label{fig:results_of_qb_IN}
		\includegraphics[width=3.2cm]{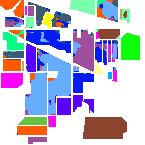}} \hspace{12pt}\vspace{2pt}
	\subfigure[MDGCN\cite{wan2019multiscale}]{\label{fig:results_of_qb_IN}
		\includegraphics[width=3.2cm]{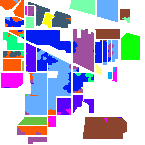}} \hspace{12pt}\vspace{2pt}
	\subfigure[RNN\cite{mou2017deep}]{\label{fig:results_of_qb_IN}
	    \hspace{12pt} \includegraphics[width=3.2cm]{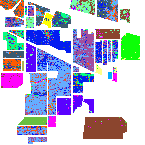}} \hspace{12pt}\vspace{2pt}
	\subfigure[SSRN\cite{zhong2017spectral}]{\label{fig:results_of_qb_IN}
		\includegraphics[width=3.2cm]{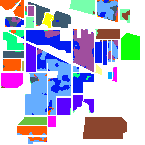}} \hspace{12pt}\vspace{2pt}
	\subfigure[FDSSN\cite{wang2018fast}]{\label{fig:results_of_qb_IN}
		\includegraphics[width=3.2cm]{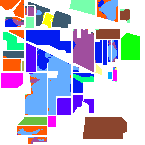}} \hspace{12pt}\vspace{2pt}
	\subfigure[Proposed]{\label{fig:results_of_qb_IN}
		\includegraphics[width=3.2cm]{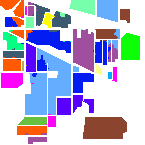}} \hspace{12pt}\vspace{2pt}
\caption{Prediction results by different methods on Indian Pines data set.}
\label{indian exp pic}
\end{figure*}

\begin{figure*}[!htb]\vspace{10pt}
	\subfigure[Ground-truth]{\label{fig:results_of_qb_UP}
		\includegraphics[width=3.6cm]{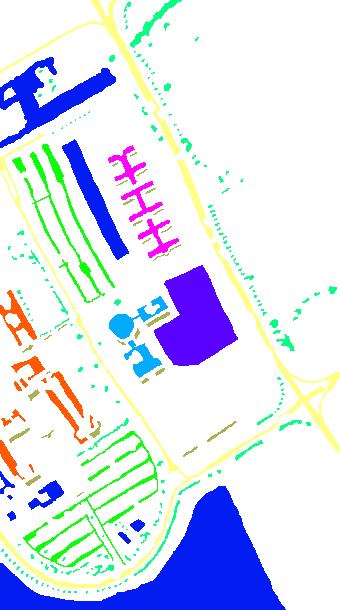}} \hspace{2pt}\vspace{0pt}
	\subfigure[EMAP\cite{plaza2004new}]{\label{fig:results_of_qb_UP}
		\includegraphics[width=3.6cm]{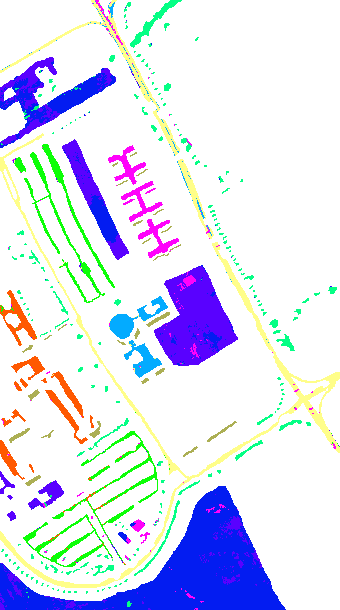}} \hspace{2pt}\vspace{0pt}
	\subfigure[PResNet\cite{paoletti2018deep}]{\label{fig:results_of_qb_UP}
		\includegraphics[width=3.6cm]{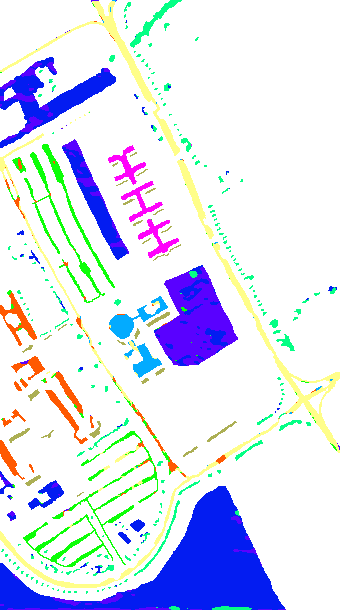}} \hspace{2pt}\vspace{0pt}
	\subfigure[MDGCN\cite{wan2019multiscale}]{\label{fig:results_of_qb_UP}
		\includegraphics[width=3.6cm]{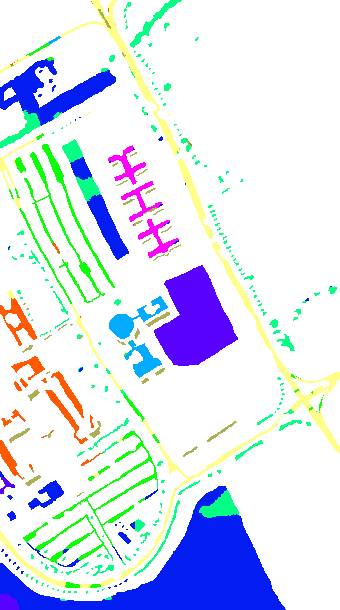}} \hspace{2pt}\vspace{0pt}
	\subfigure[RNN\cite{mou2017deep}]{\label{fig:results_of_qb_UP}
		\includegraphics[width=3.6cm]{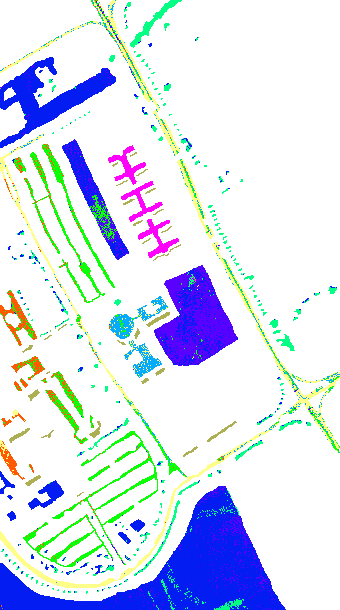}} \hspace{20pt}\vspace{0pt}
	\subfigure[SSRN\cite{zhong2017spectral}]{\label{fig:results_of_qb_UP}
		\includegraphics[width=3.6cm]{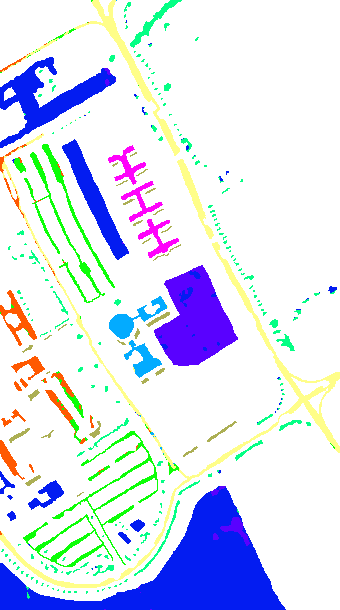}} \hspace{20pt}\vspace{0pt}
	\subfigure[FDSSN\cite{wang2018fast}]{\label{fig:results_of_qb_UP}
		\includegraphics[width=3.6cm]{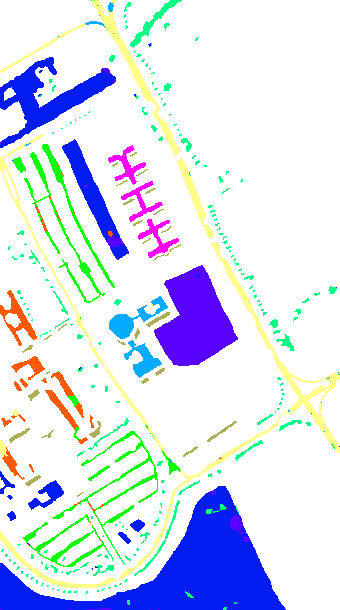}} \hspace{20pt}\vspace{0pt}
	\subfigure[Proposed]{\label{fig:results_of_qb_UP}
		\includegraphics[width=3.6cm]{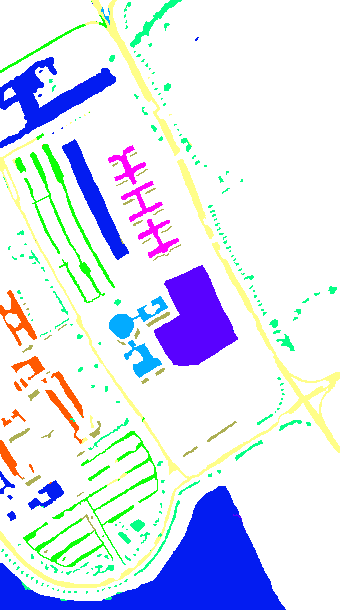}} \hspace{2pt}\vspace{0pt}
\caption{Prediction results by different methods on University of Pavia data set.}
\label{pavia exp pic}
\end{figure*}

Our workflow is kept identical for all the datasets.
The number of hidden units is fixed as 64; the number of steps for each epoch is five times as much as number of clusters; and the number of epochs and the learning rate are set to 400 and 0.005, respectively. 
We use ADAM \cite{kingma2014adam} to optimize our model.
For each reported result, the experimental trial is repeated ten times and the average and standard deviation are used for measurement.

\subsection{Comparison with other methods}
We first compare the classification accuracy of our proposed framework with the aforementioned state-of-the-art methods. 
Note, we do not conduct the experiment with S$^2$GCN due to its huge memory consumption, and instead, we cite the results of Indian Pines and University of Pavia datasets from \cite{wan2019multiscale}.

The classification results on Indian Pines data are shown in Fig. \ref{indian exp pic} and Table \ref{indian exp tab}. 
It can be observed that RNN gives poor performance on this dataset, which is mostly due to the complex spectral distribution of the land-cover.
Compared with other methods, our method achieves the best performance on the three performance metrics, and even 100\% accuracy for class 1, 7, 8, 9, 12. 
Visualization results are shown in Fig. \ref{indian exp pic}, illustrating that our method shows the best spatial consistency compared to other baseline methods.

\begin{figure*}[!htb]
	\subfigure[Ground-truth]{\label{fig:results_of_qb_SA}
		\includegraphics[width=3.6cm]{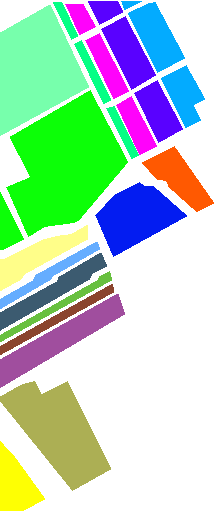}} \hspace{2pt}
	\subfigure[EMAP\cite{plaza2004new}]{\label{fig:results_of_qb_SA}
		\includegraphics[width=3.6cm]{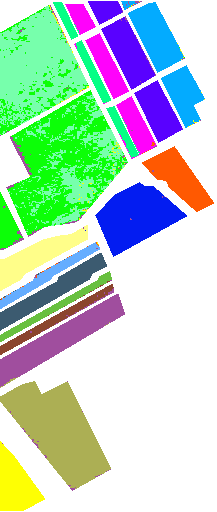}} \hspace{2pt}
	\subfigure[PResNet\cite{paoletti2018deep}]{\label{fig:results_of_qb_SA}
		\includegraphics[width=3.6cm]{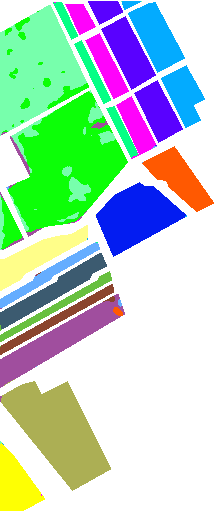}} \hspace{2pt}
	\subfigure[MDGCN\cite{wan2019multiscale}]{\label{fig:results_of_qb_SA}
		\includegraphics[width=3.6cm]{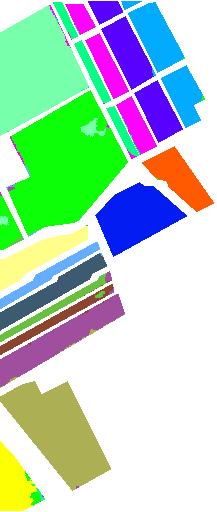}} \hspace{2pt}
	\subfigure[RNN\cite{mou2017deep}]{\label{fig:results_of_qb_SA}
		\includegraphics[width=3.6cm]{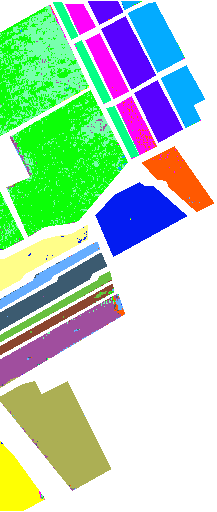}} \hspace{20pt}
	\subfigure[SSRN\cite{zhong2017spectral}]{\label{fig:results_of_qb_SA}
		\includegraphics[width=3.6cm]{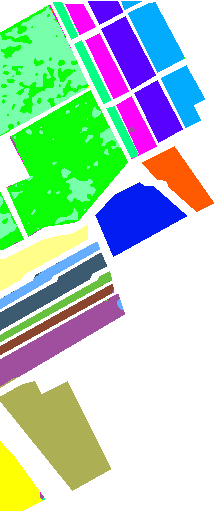}} \hspace{20pt}
	\subfigure[FDSSN\cite{wang2018fast}]{\label{fig:results_of_qb_SA}
		\includegraphics[width=3.6cm]{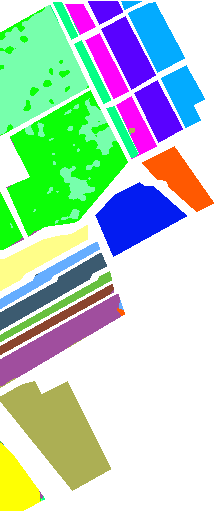}} \hspace{20pt}
	\subfigure[Proposed]{\label{fig:results_of_qb_SA}
		\includegraphics[width=3.6cm]{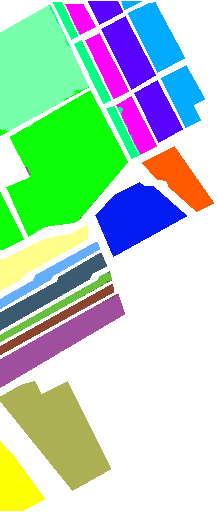}} \hspace{2pt}
\caption{Prediction results by different methods on Salinas dataset. }
\label{salinas exp pic}
\end{figure*}

\begin{figure*}[!htb]
	\subfigure[IP]{\label{fig:results_of_qb_SA}
		\includegraphics[width=5.7cm]{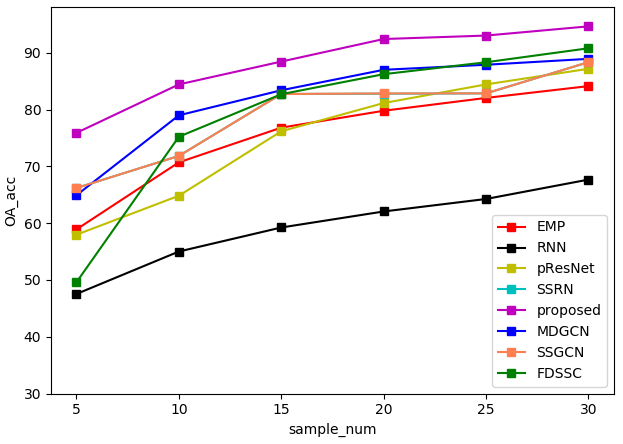}} \hspace{2pt}\vspace{0pt}
	\subfigure[UP]{\label{fig:results_of_qb_SA}
		\includegraphics[width=5.7cm]{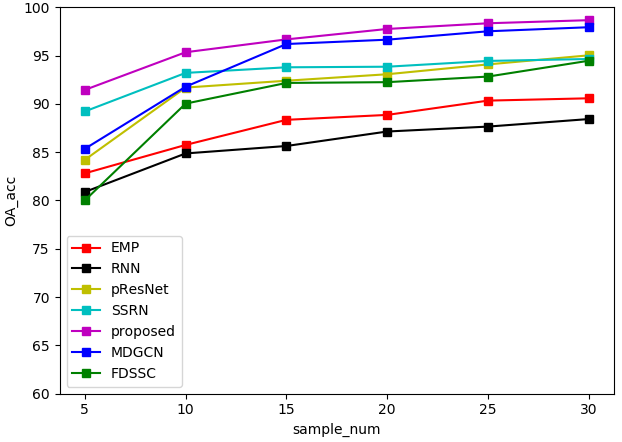}} \hspace{2pt}\vspace{0pt}
	\subfigure[SA]{\label{fig:results_of_qb_SA}
		\includegraphics[width=5.7cm]{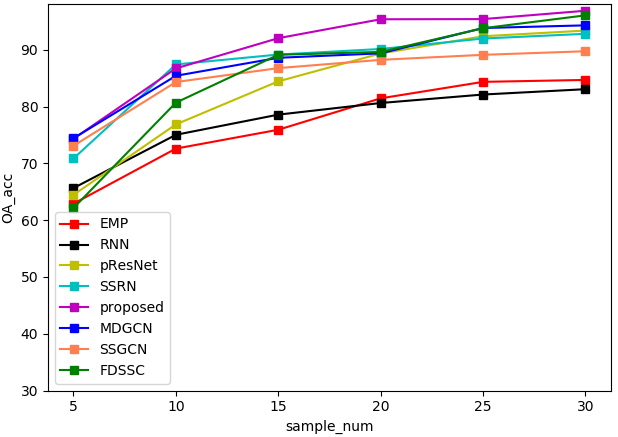}} \hspace{2pt}\vspace{0pt}
\caption{OA accuracies under different numbers of labeled examples per class. Left: Indian Pines dataset. Middle: University of Pavia dataset. Right: Salinas dataset. }
\label{ab_study:numbers}
\end{figure*}

The results on the University of Pavia dataset are shown in Table \ref{pavia exp tab}.
It can be seen that our method achieves the best OA accuracy, i.e. 96.87\%, when only 30 training samples of each class are provided. 
Note that FDSSC also gets a promising OA accuracy 96.06\%.
However, its AA accuracy is 2.39\% lower than ours.
Fig. \ref{pavia exp pic} illustrates the classification results of the compared methods on University of Pavia data, from which it can be observed that our method generates more homogeneous labels in terms of spatial features.

Table~\ref{salinas exp tab} presents the experimental results of different methods on the Salinas data set. 
It can be observed that the performance of all compared methods is better than that on Indian Pines and University of Pavia data sets.
The main reason may be that the land-cover of Salinas data set is relatively centralized and the spectral diversity is low, which is conductive to classification. 
Our method still achieves the best performance in terms of the three benchmark metrics. 
As shown in Fig. \ref{salinas exp pic}, it can be found that our method gains remarkable accuracy on edges and chunked area of land-cover.

\subsection{Performance regarding number of training samples}

\begin{figure*}[!htb]\vspace{0pt}
	\subfigure[IP]{\label{fig:results_of_qb_SA}
		\includegraphics[width=5.7cm]{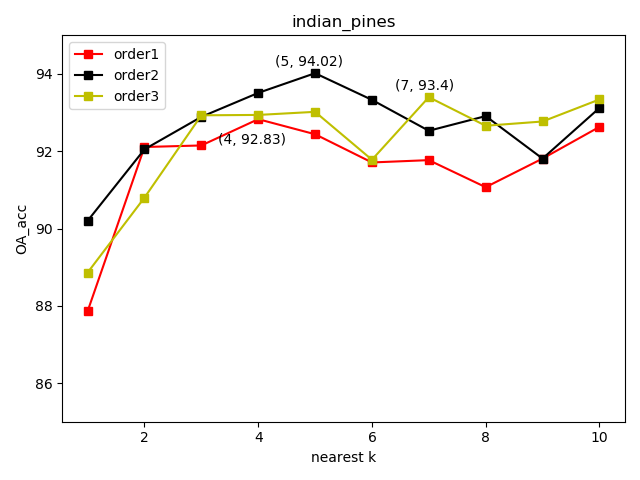}} \hspace{2pt}\vspace{0pt}
	\subfigure[UP]{\label{fig:results_of_qb_SA}
		\includegraphics[width=5.7cm]{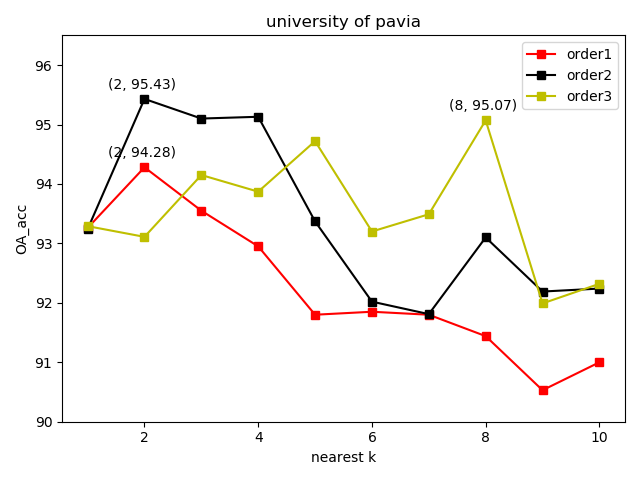}} \hspace{2pt}\vspace{0pt}
	\subfigure[SA]{\label{fig:results_of_qb_SA}
		\includegraphics[width=5.7cm]{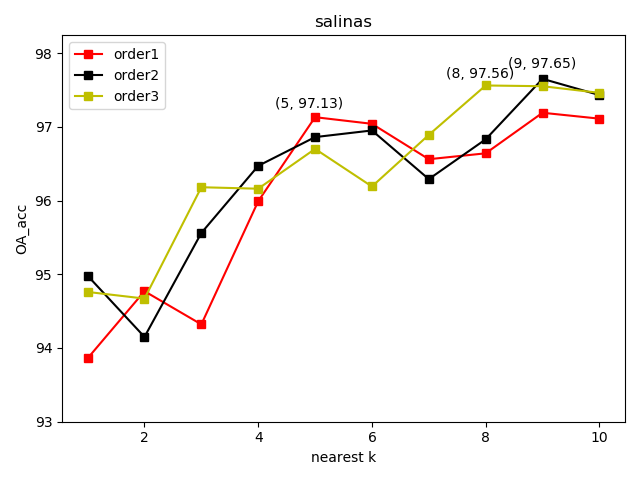}} \hspace{2pt}\vspace{0pt}
\caption{OA accuracies of proposed method regarding hyper-parameters $o$ and $k$. Left: Indian Pines. Middle: University of Pavia. Right: Salinas.}
\label{ab_study:hyper_parameter_o_k}
\end{figure*}

\begin{table*}[!h]
\renewcommand\arraystretch{1.3}
\centering
\caption{OA ACCURACIES REGARDING DIFFERENT NUMBERS OF GRAPH PARTITIONS}\label{tab:aStrangeTable}
\vspace{0.1cm}
\begin{tabular}{ccccccccccc}
\toprule
Dataset& $c$ = 1& $c$ = 3& $c$ = 5& $c$ = 7& $c$ = 9& $c$ = 11& $c$ = 13& $c$ = 15& MAX\\
\midrule
IP& \textbf{94.02$\pm$1.24}& 93.23$\pm$1.00& \textbf{94.65$\pm$1.21}& 94.21$\pm$1.38& 94.27$\pm$0.8& 94.11$\pm$1.19& 93.40$\pm$0.91& 94.33$\pm$1.12&  0.63\\
UP& \textbf{95.43$\pm$2.42}& 95.51$\pm$1.76& 96.47$\pm$1.51& \textbf{96.87$\pm$1.11}& 95.83$\pm$1.96& 95.66$\pm$1.99& 96.48$\pm$1.04& 96.46$\pm$0.57& 1.44\\
SA& \textbf{97.65$\pm$1.05}& 98.03$\pm$1.01& \textbf{98.67$\pm$0.60}& 98.50$\pm$0.56& 98.42$\pm$0.58& 98.24$\pm$0.67& 98.45$\pm$0.70& 98.05$\pm$0.84& 1.02\\
\bottomrule
\end{tabular}
\label{ab_study:hyper_parameter_p}
\end{table*}

In this group of experiments, we explore the performance of our method and other competitors when various numbers of training examples are given.
We train each method with the number of samples that is set to be 5, 10, 15, 20, 25, 30 on each dataset.
Results are shown in Fig. \ref{ab_study:numbers}. 
It can be seen that more training samples can yield higher accuracy for every method in the experiments. 
Generally, we can see that our method outperforms other baseline methods in most training sample number settings on the three dataset, with only one exception, i.e. when 10 training samples provided on Salinas dataset. 
We also observe that, compared with other convolution methods except RNN, the superpixel-based methods, i.e. the proposed method and MDGCN, can gain comparably high accuracy provided with fewer samples. 
The accuracy of our method is higher than that of MDGCN.
RNN can achieve better performance than MDGCN when only 5 training samples are provided, especially on Pavia dataset.
However, its performance can not get further enhanced as the number of samples increases.

\begin{table*}[!h]
\renewcommand\arraystretch{1.3}
\centering
\caption{TEST TIME (IN SECONDS) OF DEEP METHODS ON THREE DATASETS}\label{tab:aStrangeTable}
\vspace{0.1cm}
\begin{tabular}{ccccccc}
\toprule
Dataset& pResNet\cite{paoletti2018deep}& MDGCN \cite{wan2019multiscale}&  FDSSC\cite{wang2018fast}& RNN\cite{mou2017deep}& SSRN\cite{zhong2017spectral}& proposed\\
\midrule
IP& 2.09& 109.12& 3.27& \textbf{0.31}& 4.00& 0.54\\
UP& 4.23& 445.84& 11.50& 1.44& 18.17& \textbf{0.60}\\
SA& 7.99& 190.93& 18.43& 1.66& 21.63& \textbf{0.55}\\
\bottomrule
\end{tabular}
\label{runtime}
\end{table*}

\subsection{Ablation study}
We then examine the performance of our method regarding the three hyper-parameters in our algorithm, including  1) $o$ in Eq.~(\ref{eq2}) in Sec.~\ref{graph construction}, which represents the scale of the order of nodes covered in our adjacency matrix;
2) $k$ in Eq.~(\ref{eq1}) in Sec.~\ref{graph construction}, which represents the top-k highest similarities in multi-scale neighbor nodes we select;
and 3) $p$ which is the number of sub-graphs after graph partition in Sec.~\ref{graph partition}. 

$ \bullet $  $ \textit{o and k} $: Both hyperparameters are devoted to contributing to graph construction. 
We thus design an experiment setting for excavating a good pair of $o$ and $k$ from $o$ is set in a range from 1 to 3, the value of $k$ is set in a range from 1 to 10, respectively. 
In order to eliminate the impact of the second clustering and number of training samples, the number of partitions and number of training samples are fixed as 1 and 30 respectively, which means the second clustering (i.e. graph partition) will not be applied.
The results on the three datasets are shown in Fig. \ref{ab_study:hyper_parameter_o_k}.
It can be seen that the best result is always achieved when $o$ is set as 2, and when $o$ is set as 1 which means multi-scale adjacency is not considered, the performance is worse than that when $o$ is equal to 2 or 3.
These results prove that a multi-scales adjacency matrix indeed brings better performance.

From Fig. \ref{ab_study:hyper_parameter_o_k}, we can find the highest accuracy is achieved when $k$ is set as 5, 2, 9, respectively for Indian Pines, University of Pavia and Salinas datasets.  
After observing the ground-truths of three datasets, we find the size of continuous homogeneous regions for each dataset is different.
For the University of Pavia dataset, due to the scattered buildings, its regional continuous homogeneity shows irregular shapes, which means the adjacency of each superpixel is very complex, and the small $k$ (2 in here) is more suitable for finding the reliable relations between a node and its adjacent nodes. 
For the Indian Pines datatset and Salinas dataset, the distributions of land-cover are more continuous and regular, and the corresponding $k$ achieves the best performance on a relatively large $k$ value which are 5 and  9, respectively. 
Such results fit the common belief that more neighbors information would be useful for classification over a large and coherent HSI.

$ \bullet $ $ \textit{c} $: We conduct experiments with varied values of $c$, namely the number of sub-graphs defined in graph partition of our method, to explore the effects of the graph partition operation. 
We check the performance of our proposed framework by changing the value of $c$ from 1 (meaning never applying the second clustering on the superpixel graph) to 15.
The results are shown in Table \ref{ab_study:hyper_parameter_p}. 
The hyper-parameters of graph construction are set the same as the best settings, i.e. $o = 2$, $k = 5, 2, 9$ respectively for Indian Pines, University of Pavia, and Salinas datasets.
The results show that the second clustering, i.e. the graph partition operation, improves the classification performance in most cases except when $ c $ is equal to 3 and 13 on Indian Pines dataset.
This is possibly because the second clustering not only reduces the weak dependencies but also cuts some meaningful edges sometimes. 
The performance on the University of Pavia and Salinas datasets is always higher than that when no second clustering is applied.
In particular, the second clustering brings 1.44\% and 1.02\%  when $ c $ is 7 and 5 respectively.

\subsection{Run time}
In this subsection, we evaluate the run time of our proposed method.  
Table~\uppercase\expandafter{\romannumeral8} shows the test time of baseline models including MDGCN, pResNet, FDSSC, RNN, SSRN and our proposed method on the three benchmark datasets.
All the test experiments are conducted on a 2.2-GHz Intel Xeon CPU and a GeForce RTX 2020ti GPU.
As shown in Table~\ref{runtime}, we find that the bigger a dataset is, the more run time a tested model will cost, except for the superpixels-based methods MDGCN and our method. 
Our method uses the least run time on the University of Pavia and Salinas datasets.
On Indian Pines dataset, the run time of RNN is shorter than our method, but note that this dataset is relatively small. Our model has good time performance on large-scale datasets.
Compared with MDGCN, which is also superpixel based graph method, our method achieves less time consumption based on the same superpixels graph due to our simpler model structure and double clustering operations.

\section{Conclusion}
In this work, we have proposed a novel semi-supervised GCN framework for HSI classification.
Given an input HSI, we first transform it into a superpixel level graph to relieve the huge computational consumption.
Before applying GCN to classification, we further partition this graph into several sub-graphs, in which process we remove unimportant edges to strengthen those important ones.
With such a dual-clustering operation, our framework can better exploit the multi-hop node correlations and effectively reduce the computation burden. 
Experimental results on three widely used hyperspectral image datasets demonstrate that the proposed framework is able to yield better performance when comparing with state-of-the-art methods.


%









\bibliographystyle{./references/IEEEtran}
\bibliography{./references/IEEEabrv,./references/refs_dataset}
\end{document}